# Personalized Classifier for Food Image Recognition

Shota Horiguchi, *Member, IEEE,* Sosuke Amano, Makoto Ogawa, and Kiyoharu Aizawa, *Fellow, IEEE.*

*Abstract*—Currently, food image recognition tasks are evaluated against fixed datasets. However, in real-world conditions, there are cases in which the number of samples in each class continues to increase and samples from novel classes appear. In particular, dynamic datasets in which each individual user creates samples and continues the updating process often have content that varies considerably between different users, and the number of samples per person is very limited. A single classifier common to all users cannot handle such dynamic data. Bridging the gap between the laboratory environment and the real world has not yet been accomplished on a large scale. Personalizing a classifier incrementally for each user is a promising way to do this. In this paper, we address the personalization problem, which involves adapting to the user's domain incrementally using a very limited number of samples. We propose a simple yet effective personalization framework which is a combination of the nearest class mean classifier and the 1-nearest neighbor classifier based on deep features. To conduct realistic experiments, we made use of a new dataset of daily food images collected by a food-logging application. Experimental results show that our proposed method significantly outperforms existing methods.

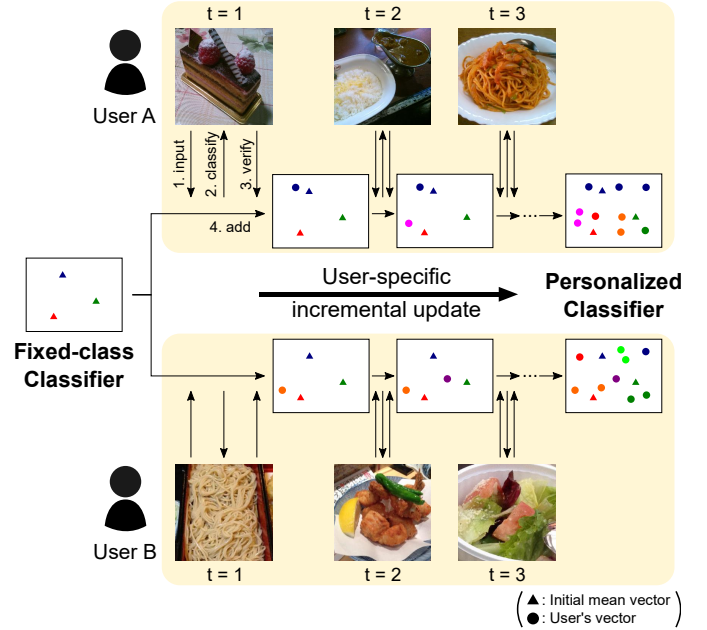

Fig. 1. Overview of the proposed sequential personalized classifier (SPC). Each user has a common nearest class mean classifier that attains competitive accuracy with a fixed-class CNN-based classifier. The classifier is gradually personalized by learning new samples incrementally from existing or novel classes using a very limited number of samples.

## I. INTRODUCTION

IMAGE classification using convolutional neural networks (CNNs) is being widely developed for many different purposes. CNNs show state-of-the-art performance in fixed-class image recognition tasks using closed datasets like ImageNet [1]. When such classifiers are used in actual scenarios, e.g., smartphone applications, classes included in the dataset do not match for all users. This causes a discrepancy between the laboratory environment and real-world situations. To bridge the gap between controlled scenarios and the real world, personalization of classifiers should be considered. Previous studies have shown that personalization can improve image annotation performance [2], [3]. However, these studies assumed a "large vocabulary" containing all the tags appearing in their experiments, whereas in actual situations, novel classes always appear. In the real world, a high diversity of classes cannot be completely accounted for at the beginning.

In this paper, we deal with such real-world problems by personalizing classifiers sequentially. Incremental learning, in which additional samples of existing classes or novel classes are learned without retraining, is necessary for obtaining visual knowledge from sequential data. Conventional methods learn metrics to classify existing classes using a large number of samples at first, and then learn novel classes incrementally. However, most of them retain constraint assumptions [4], [5], [6]. One is that the number of samples of the initial classes does not increase. In actual situations, the domains of the initial classes may change. Another is that newly trained classes have nearly the same number of samples as initially prepared classes. In the real world, novel classes have a much smaller number of samples than initial classes. Therefore, existing incremental learning techniques are insufficient for personalization.

In addition to incremental learning, personalization of classifiers should also fulfill two other aspects. One is domain adaptation. For each person, the domain, that is, the data comprised in each class, is different. The class definitions are different; thus, it is important to assume that each personal data item is from a different domain. The other is one-shot learning. When focusing on a given person, the number of images is limited; thus, it is also an important factor in personalization that new classes can be learned by using only one or a few samples.

Food image classification is one practical use case of image recognition technology. There are many studies in food image recognition [7], [8], [9], [10], [11], [12], [13], [14], [15].

S. Horiguchi, S. Amano, and K. Aizawa are with the Department of Information and Communication Engineering, The University of Tokyo, Tokyo, Japan (e-mail:horiguchi@hal.t.u-tokyo.ac.jp, s_amano@hal.t.u-tokyo.ac.jp, aizawa@hal.t.u-tokyo.ac.jp).
M. Ogawa is with foo.log Inc., Tokyo, Japan (e-mail:ogawa@foo-log.co.jp).
Manuscript received August 6, 2017; revised November 14, 2017 and January 27, 2018.

Almost all of the previous studies followed the general method of using fixed food image datasets. However, such a controlled scenario is not appropriate for a real-world purpose. We have been running a food-logging application for smartphones and have built a food image dataset named FLD consisting of over 1.5 million images. They are food images uploaded daily by the users, and as a result, the number of classes grew incrementally. Analysis of FLD showed that the appearance of food images differs considerably between people. There is high intra-class diversity and high inter-class similarity among people (see Section III).

To deal with such dynamic data, we propose a simple yet powerful personalization framework called the sequential personalized classifier (SPC), as shown in Fig. 1. We propose a personalized classifier based on the weighted nearest neighbor method, which attains performance comparable to that of the CNN classifier at first and becomes more accurate after sequential personalization without any heavy retraining. We think the simplicity of our method is very important for the personalization task in the un-controlled environment. Because of the simplicity of the weighted nearest neighbor, addition and removal of the user specific data is easily handled without any training. An important result of this paper is that the simple method outperformed more complex latest methodologies.

Our contributions are summarized as follows:

- We propose a simple yet powerful personalization framework called the sequential personalized classifier (SPC). This attains performance comparable to that of the CNN classifier at first and becomes more accurate after sequential personalization without any heavy retraining.
- We evaluated SPC's real-world performance on a new dataset of daily food images we built to reproduce real situations, rather than artificially designed simulations of real-world data [5], [6], [16], [4].
- We conducted exhaustive comparisons against previous methods and found that the proposed method significantly outperforms existing techniques.

## II. Related work

### A. The personalization problem in image recognition

Most of today's computer vision methods implicitly assume that all the classes and variations within a class are given in the training images. However, considering a more realistic scenario, only a limited number of training samples or classes are available initially; thus, incremental learning is necessary. In addition, especially in personalization problems, the number of samples from the target domain is very limited. Therefore, domain adaptation and one-shot learning should be considered as well. In this section, we discuss these three concepts related to the personalization problem. We also mention personalization methods used for tag prediction from images uploaded to photo-sharing web sites.

*1) Incremental learning:* Incremental learning assumes that the number of samples or classes is limited initially and that additional samples arrive sequentially. There are several incremental learning methods: methods based on support vector machines (SVM) [17], [18], [19]; nearest neighbor (NN) or nearest class mean (NCM) methods [20], [5], [21]; and random forests [6]. SVM-based methods suffer from the high retraining costs required to increase the number of classes. Methods based on random forests take less training time than those based on NCM+metric learning or SVM. However, only class-wise incremental learning was considered in [6]; sample-wise incremental learning for methods based on random forests still requires long training times. The NN and NCM methods are the most scalable because they can add samples or classes at almost zero cost. However, almost all incremental learning experiments assume that new classes have approximately the same number of samples as the initial classes. They also have difficulty in dealing with changes in class definitions because they treat pre-entered vectors and subsequently added samples equally. Another problem is that their methods use handcrafted features, which show considerably lower accuracy than recent CNN-based methods.

*2) Domain adaptation:* Domain adaptation is needed when there are variations in the content of classes between source and target domains. There are many domain adaptation methods [22], [23], [24], but most of them cannot learn incrementally. [25] and [26] proposed an incremental domain adaptation method for gradually changing domains, but this method does not allow addition of new classes. [16] is similar to our paper in its assumption that test images arrive sequentially; however, its experimental settings are artificial, and the number of classes did not change during the experiments.

Incremental domain adaptation for detection tasks has been studied mainly in video recognition, e.g., face detection [27] and pedestrian detection [28], [29], [30]. [31] proposed an image-to-video multi-class incremental domain expansion method, but their method cannot learn new classes.

*3) One-shot learning:* One-shot learning aims to learn new classes using only a few samples. [32], [33], and [34] conducted their experiments using local or patch-based features. [35] used deep features extracted from pre-trained networks for one-shot scene recognition using a dataset of sequentially changing outdoor scene images. In recent years, deep-learning-based methods have appeared [36], [37], [38].

All these methods encounter the problem that their evaluation methods ignore initial classes when evaluating one-shot learning performance. For the personalization problem, performance on initial classes having rich numbers of samples should be considered as well.

*4) Personalized image annotation:* Personalized image annotation is studied mainly in tag prediction using image datasets such as the NUS-WIDE dataset [39], which is crawled from Flickr, and the Facebook dataset [40]. Previous studies [2], [3] succeeded in learning user preferences from only a few samples, but they assumed that a large vocabulary was known beforehand, which does not hold in real-world situations. It is also problematic to apply their methods to personalized classification because tag prediction is a multi-label problem.

### B. Food image recognition

There have been many studies on food image recognition: for classification and detection [7], [9], [10], [11], [41], [42];

for volume or calorie estimation [12], [13], [14]; for ingredient recognition [15]; learning concept embedding [43]; and for recipe retrieval [44], [45], [46], [47]. Personalized food image recognition and calorie estimation are addressed in [48] and [49], respectively, but their experiments were conducted on datasets with a very limited number of classes and samples with handcrafted features.

Along with research on food image recognition, several food image datasets have been released to date, such as UECFOOD100 [9], Food-101 [11], and VIREO Food-172 [15]. The properties of datasets, such as how their data are collected and what classes are defined, are important. Images in these datasets, however, were collected by anonymous photographers by querying food names in public web resources [9], [11], [15]; thus, these databases are not suited to this research investigating a personalization framework.

## III. EXPERIMENTAL DATASET

To conduct our experiments, it was necessary to prepare a large-scale image dataset that contains owner IDs and time stamps with its images. The NUS-WIDE dataset [39] has been used in personalized annotation contexts [2], [3]. However, its ground truth labels are tags, which are not only object classes but also attributes (such as "cute" or "delicious") or even unrelated words. Therefore, this database does not fit our personalization problem setting, wherein each instance in an image has one class label.

In this study, we built a food image dataset collected by running a photo-based daily food recording application [50] called FoodLog for about two years. Users use FoodLog app as follows:

1) Take a photo of dishes.
2) Select a food regions in the photo.
3) Label the region with its food name.
4) Repeat step 2 and step 3 to register all the foods in the image.

When labeling a region, the user can select from the list of candidates that the system produces. When the user cannot find any appropriate entry in them, then the user defines a new label by selecting from default label set or entering user-specific label. Our purpose is to help this labeling step by image recognition that produce highly precise candidates for the selection. After two-year operation of FoodLog app, we collected 1,508,171 food images uploaded by 20,820 users from the general public. We call this two-year dataset FLD. There are 99,314 kinds of class labels in FLD; 1,857 of them are default labels, and the remaining 97,457 are user-defined. They show a very skewed distribution, as shown in Fig. 2.

NUS-WIDE and FLD are datasets whose labels are free-form descriptions by the general public, unlike ImageNet [1]. On such datasets, it is hard to determine whether the labels defined by users are correct because there are various names for a specific object. To maintain the users' intentions, image labels should be unchanged from their original expressions and should not be standardized. In this paper, we treat all the labels in FLD as ground truth, as in the tag estimation problem using NUS-WIDE.

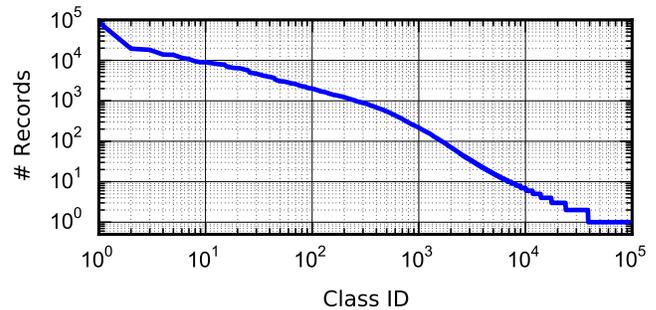

Fig. 2. Number of records per class in the FLD food image dataset.

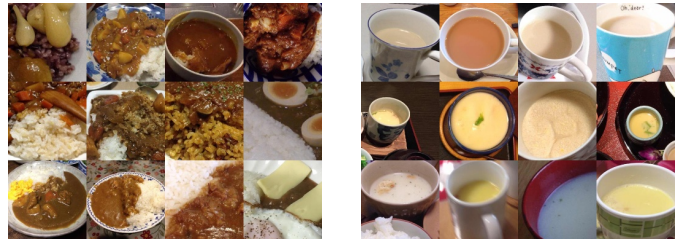

(a) Pork curry (top), beef curry (middle), and chicken curry (bottom).

(b) Milk tea (top), egg custard (middle), and instant corn cream soup (bottom).

Fig. 3. Inter-class similarity in FLD.

In food image dataset, labels sometimes have hierarchical structures. For example, the labels in VIREO Food-172 dataset [15] are grouped into eight major categories, while the labels in UECFOOD100 [9] and Food-101 [11] do not have any grouped categories. For FLD, the default labels are hierarchically grouped: 1,857 labels are grouped into 93 middle classes, and they are grouped into 17 large classes. These hierarchical groups are defined and provided by eatsmart [1], which is one of the food databases operated by companies. Thus, hierarchical classification can be provided for the default labels. However, 97,457 user-defined classes do not have hierarchical classes at all. Hierarchical information is not very trivial and it is not practical to annotate all the user-defined classes with hierarchical information.

As shown in Fig. 3 and Fig. 4, respectively, FLD includes significant inter-class similarity and intra-class diversity. In contrast to general fine-grained classification, there are some images that are not separable based on their visual appearance. On the other hand, there are classes that contain images having completely different visual appearances. This makes real-world food image classification difficult. However, by focusing on a particular user, these problems become less difficult. In the case of curry, there are 348 users who recorded pork, beef, or chicken curry images five times or more; the three curries are visually indistinguishable (Fig. 3a), but 146 of the users (42%) ate only one kind of curry. In the case of yogurt, there was appreciable visual diversity, but visual appearances were very similar within each user's collection, as shown in Fig. 4a. Therefore, incremental personal domain adaptation is necessary for image recognition in actual scenarios.

---
[1] https://www.eatsmart.jp/

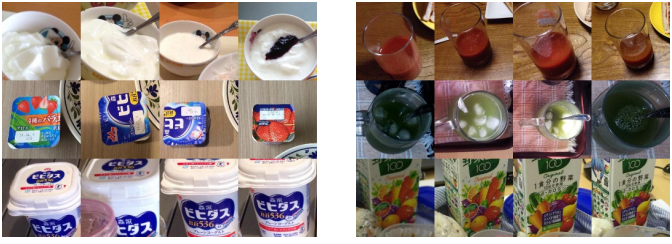

(a) Each row shows the first four yogurt records of a specific user.

(b) Each row shows the first four vegetable juice records of a specific user.

Fig. 4. Intra-class diversity in FLD.

For our experiments, we used 213 major classes from the default label set for training CNNs. Each image was cropped according to the user's annotations and resized to 256×256 pixels. For details of the method used to determine the number of classes, see Section V-B. We refer to this subset as FLD-213. We show the example from each class in Fig. 5.

Detailed information of classes in FLD-213 are described in Appendix A. They are not always exclusive: for example, both *coffee* and *black coffee* are included in FLD-213. As described in Section V, we trained a standard softmax-based classifier using them. Our purpose is to predict the single label which the user uses for the image. Multi-label approach could be applicable, but there are difficulties on making use of it. First, in order to apply a multi-label approach, we have to produce additional annotations besides the user-defined annotation. It is infeasible to assign multiple appropriate labels from hundreds of label candidates to over a million images. Second, even if we would extend our dataset to be suited for multi-label problem, evaluation of the predicted labels would be difficult — ranking the predicted labels would be difficult because we have only a single label annotated by a user per an image.

We used another subset of FLD for our personalized classification experiments; we refer to this subset as FLD-PERSONAL. This subset did not overlap with the dataset used for training CNNs. FLD-PERSONAL consisted of 276,900 images, representing the first 300 food records from 923 different users. Each image was cropped from its original whole image according to the user's annotations and was resized to 256×256 pixels.

## IV. PROPOSED METHOD

We propose a fast domain adaptation method that can learn novel classes incrementally. In the real world, the number of occurrences of classes is very biased, as shown in Fig. 2. Facebook hashtags have a similar distribution [40]. To classify such unbalanced data, it is practical to make use of a fixed-class classifier for major classes that appear frequently. In addition, we have to deal with the increasing number of classes. We propose a method that can classify frequent classes with competitive accuracy relative to the CNN classifier [51] initially and that increases in accuracy by sequentially learning the user's input. Previous methods used handcrafted features [20], [21], [6]; thus, it is not proven whether their methods attained an accuracy comparable to the CNN-based method. We reveal that personalization can outperform general methods that use one CNN for each user.

### A. Personalized object recognition using SPCs

We propose a personalization framework called a sequential personalized classifier (SPC), which can learn novel classes incrementally. First, we trained a convolutional neural network (CNN) as a fixed-class classifier. We tried various sizes of datasets and optimized the dataset size (see Section V-B for optimization details). We extracted features from the CNN and calculated the means of each class using training samples. We found that NCM classifiers, which use their mean vectors for each class, performed as well as the CNN softmax classifier. It has also been shown that deep features perform very well in other datasets [52], [53]. This high generality is beneficial to novel class learning.

Each user $u \in U$ has his/her own database $V_u$. $V_u$ is empty at time $t = 0$, and the user's records are registered into $V_u$ at each time; thus, after the user's $t$th record, $V_u$ is denoted by

$$V_u = \{(f(\mathbf{x}_{ui}), c_{ui}) | 1 \leq i \leq t\}, \quad (1)$$

where $\mathbf{x}_{ui}$ and $c_{ui}$ represent the user $u$'s $i$th record and the class to which it belongs, respectively. $f(\cdot)$ is the feature extracted from the network. Personalized classification is conducted using $V_u$ and the set of mean vectors $V_m$, which is common to all users. $V_m$ is defined as

$$V_m = \{(f(\mathbf{x}_{mi}), c_{mi}) | 1 \leq i \leq |C_m|\}, \quad (2)$$

where $C_m$ is the set of classes in $V_m$.

In our sequential personalized classification framework, when the user records the $t$th dish $\mathbf{x}_{ut}$, its class $c_{ut}^*$ is predicted by making use of both the common $V_m$ and user's own $V_u$. Similarities of $\mathbf{x}_{ut}$ to a class $c_m$ in the common $V_m$ and to a class $c_u$ in the user's own $V_u$ are computed by $s(c_m, \mathbf{x}_{ut}, V_m)$ and $s(c_u, \mathbf{x}_{ut}, V_u)$, respectively. We use the general dot product as the similarity metric:

$$s(c, \mathbf{x}_{ut}, V) = \begin{cases} \max_{(f(\mathbf{x}),c) \in V_c} f(\mathbf{x})^T f(\mathbf{x}_{ut}) & \text{if } V_c \neq \emptyset \\ 0 & \text{otherwise} \end{cases}, \quad (3)$$

where $V_c$ is a subset of $V$ whose samples belong to class $c$. In order to combine the two similarities, we introduce a parameter $w > 0$ as a weighting factor, and the class $c_{ut}^*$ is finally predicted by the following equation:

$$c_{ut}^* = \arg\max_{c \in C} \left[\max\{s(c, \mathbf{x}_{ut}, V_u), w \cdot s(c, \mathbf{x}_{ut}, V_m)\}\right]. \quad (4)$$

$C$ is the union of $C_u$ and $C_m$, where $C_u$ is the set of classes observed in $V_u$.

$w$ controls the degree of personalization, that is, the balance between the common means $V_m$ and the user's vectors $V_u$. Our method accelerates incremental domain adaptation by weighting the user's past inputs $V_u$ more heavily than common mean vectors $V_m$, which means $w$ should be set smaller than 1.0. As described in Section V, we optimize $w$ by cross-validation.

We also tried a combination of the two similarities as follows instead of Eq. (4):

$$c_{ut}^* = \arg\max_{c \in C} \left[(1 - w_s) \cdot s(c, \mathbf{x}_{ut}, V_u) + w_s \cdot s(c, \mathbf{x}_{ut}, V_m)\right], \quad (5)$$

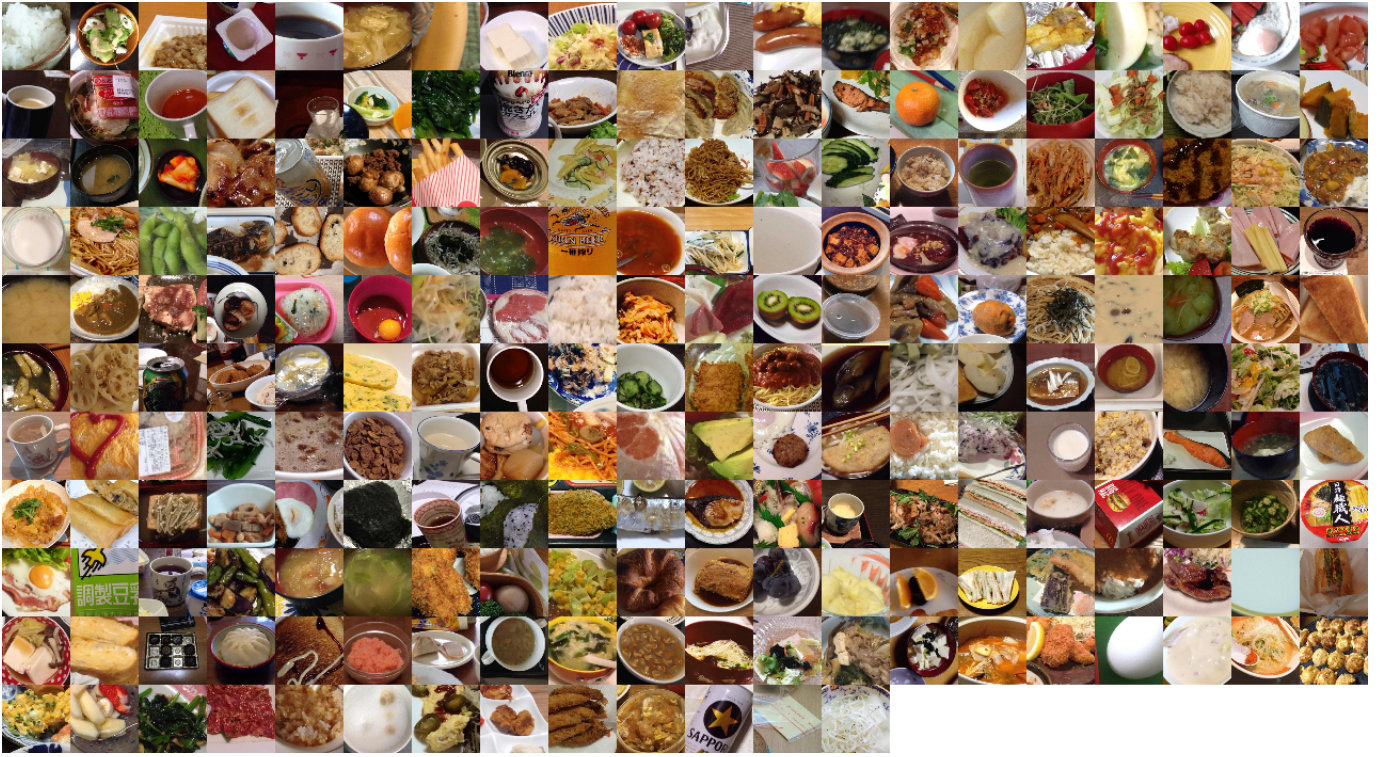

Fig. 5. Examples from FLD-213.

where $w_s$ is a balancing parameter between the two similarity. However, this did not work well. For example, assuming the input class $c$ is common to both $C_u$ and $C_m$, the combination is not maximized when the vector belongs to $C_u$ is not close to the vectors belongs to $C_m$. It is also a problem when the input class belongs to either $C_u$ or $C_m$—it may not obtain high similarity because of the combination. Detailed experimental results are shown in Appendix B.

In Fig. 6, we illustrate three typical cases that can cause our incremental personalization to change the original classifier. When an entered vector is far from every initial vector, its visual concept is novel. Thus, weighting the user's vector more heavily is effective because the CNN is trained without images of such a novel visual concept (Fig. 6(i)). When an entered vector is a new sample for existing classes, weighting is also effective because the new vector covers similar initial vectors (Fig. 6(ii)). This solves intra-class diversity problems such as Fig. 3a. When a user defines a new class to describe a visual concept that is the same as one of the initial classes, weighting is also important for robust concept transfer (Fig. 6(iii)). In Fig. 6(ii) and (iii), the common vectors covered by the user's vector are "forgotten," but those not covered are not affected. This local adaptation performs well for personalization, in contrast to global adaptation [30]. Testing time is also reduced by using NCM for initial classifiers that use an enormous number of samples and using NN as a user-specific classifier, wherein the number of samples is limited.

After prediction, the user corrects the predicted result if it is wrong and then adds $(f(\mathbf{x}_{ut}), c_{ut})$ to $V_u$.

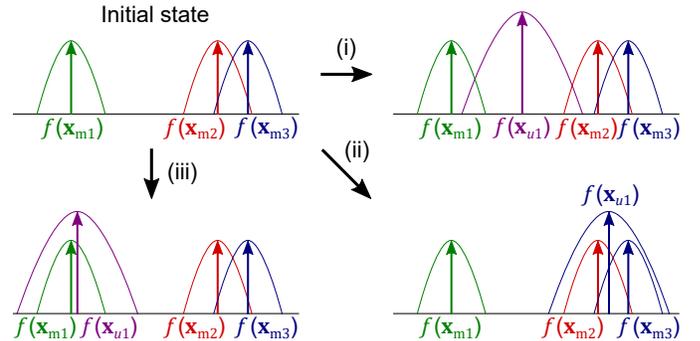

Fig. 6. The efficiency of weighting user's inputs $V_u$ more heavily than common mean vectors $V_m$. In this case, $V_u = \{f(\mathbf{x}_{u1})\}$ and $V_m = \{f(\mathbf{x}_{m1}), f(\mathbf{x}_{m2}), f(\mathbf{x}_{m3})\}$. Each arrow represents a feature vector, and its length and color represent the its weight $w$ and the class to which it belongs, respectively. The curve over each vector shows the weighted cosine similarity from the vector. Assume that $\mathbf{x}_{m1}$, $\mathbf{x}_{m2}$, and $\mathbf{x}_{m3}$ belong to the "orange," "beef curry," and "pork curry" classes, respectively. There are three typical cases in which incremental personalization changes the initial classifier: (i) When the user's sample $\mathbf{x}_{u1}$ has a visual appearance different from the initial classes. (ii) When $\mathbf{x}_{u1}$ belongs to "pork curry"; in this case, it becomes dominant for $\mathbf{x}_{m2}$ and $\mathbf{x}_{m3}$. (iii) When $\mathbf{x}_{u1}$ belongs to "tangerine"; in this case, the class "orange" is renamed to "tangerine."

## V. EXPERIMENTS

### A. Evaluation protocols

To evaluate the personalized food recognition performance of the classifier, we calculated the mean accuracy for all users $U$:

$$MeanAccuracy(t) = \frac{1}{|U|} \sum_{u \in U} \mathbb{1}\left(c_{ut}^* = c_{ut}\right), \qquad (6)$$

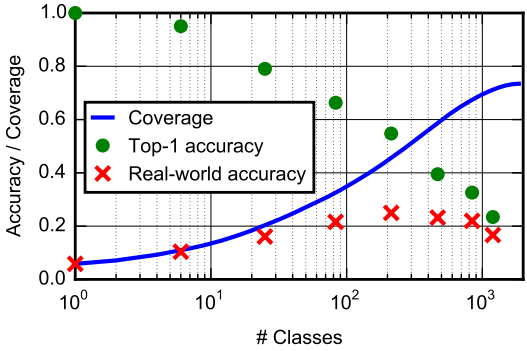

Fig. 7. Top-1 accuracy, coverage, and estimated real-world accuracy on the subsets of FLD.

where $\mathbb{1}(\cdot)$ is the indicator function. We show the top-$\{1, 5\}$ accuracy for personalized food recognition. In this study, we used FLD-PERSONAL to evaluate personalized classification performance; thus $|U| = 923$.

### B. Personalized food recognition

We evaluated our personalized food recognition method on FLD-PERSONAL. We first defined the number of initial classes as described below (Section V-B1) and tested various features extracted from the CNN as described further below (Section V-B2) before conducting our main experiments.

*1) Initial class definition:* To define the optimal number of initial classes, we trained GoogLeNet [54] with batch normalization [51] for various sizes of datasets. We fine-tuned the networks from weights pre-trained on ImageNet [1]. We selected food items from the default label set which had at least $\{100, 200, 500, 1{,}000, 2{,}000, 5{,}000, 10{,}000\}$ records. Each subset had $\{1{,}196, 841, 469, 213, 83, 25, 6\}$ classes. We randomly sampled images to have the same number of images for each class; thus, each subset had $\{119{,}600, 168{,}200, 234{,}500, 213{,}000, 166{,}000, 125{,}000, 60{,}000\}$ images. We used 80% of each subset for training and the rest for testing.

The green dots in Fig. 7 show the top-1 accuracies within the subsets; they decline almost linearly relative to the logarithm of the number of classes. However, these accuracies do not hold in the real world because many of the classes are not included in the subsets. Therefore, we roughly estimated real-world accuracies by multiplying the accuracy within the subsets by their coverage on the whole dataset (red crosses in Fig. 7). The coverage was computed by the following equation:

$$Coverage(|C_\mathrm{m}|) = \frac{\text{\#Images belonging to } C_\mathrm{m}}{\text{\#Images in FLD}}, \quad (7)$$

where $C_\mathrm{m}$ is the set of classes in the subset. This result indicates that accuracy in the real world is not high if coverage is too small or too large. We must choose the proper number of classes for the fixed-class classifier so that it is most effective in the real world. We decided to use 213 classes, which was estimated to show the highest accuracy as an initial common classifier.

*2) Feature selection:* For personalized classification, we must take into account both fixed-class and incremental classification. Fixed-class classification, which is classification within defined classes, is important for avoiding the cold-start problem. Incremental classification, whose classifier incrementally learns novel classes that are not defined in advance, is also important in measuring personalized classification performance. Therefore, in our personalized classification, it is desirable for features to attain (i) high fixed-class classification accuracy, that is, 213-class classification accuracy, and (ii) high incremental classification accuracy.

We evaluated features extracted from a classification-based network and those from a network based on distance metric learning (DML). For a classification-based network, we trained a 213-way softmax classifier using GoogLeNet with batch normalization [51] and extracted features from the *pool5* and *classifier* layers. We normalized these features to have a unit L2 norm; normalization is an important operation for gaining higher accuracy. The features of *prob* are the output of a CNN whose elements describe the probabilities of classes. For a DML-based network, we examined lifted structured feature embedding [55], which is a state-of-the-art DML method.

First, we evaluated initial accuracy using FLD-213. The results are shown in the "Fixed-class" column of Table I. For *pool5*, *classifier*, and *lifted*, we used the nearest class mean (NCM) classifier. Each centroid was computed from the features of the training set. The centroids of *pool5* and *classifier* were also normalized to have L2 unit norms. It is not surprising that *prob* features showed the best performance, but we found that the NCM classifier of *pool5* showed accuracy competitive with the *prob* features without any additional metric learning, dropping only 2.0% for top-1 and 1.2% for top-5 accuracy relative to the *prob* features.

We also estimated fully personalized performance on FLD-PERSONAL using a naive 1-NN classifier. In this case, there were no defined classes at first; thus, the classification accuracy for $t_1$ was zero. We show the average mean accuracy for $t_{251}$–$t_{300}$ for all users of 1-NN performance in the "Incremental" column of Table I. We found that *pool5* features showed the best performance. This result indicates that features extracted from a CNN have high expressiveness, which coincides with the conclusion from a previous study that features from the middle layer of a CNN trained on ImageNet can be repurposed to other classification tasks [52].

In addition, we compared *pool5* and *classifier* features from pre-trained model without fine-tuning, but in both fixed-class and incremental settings they performed worse than those from fine-tuned network. We also found that *lifted* showed worse performance in both cases than the softmax-based classifier even though it learned feature descriptors directly.

To obtain image features that are useful for both fixed-class and incremental classification, we decided to use *pool5* features for our personalization.

*3) Personalized classification:* Our final purpose is to obtain a classifier that attains high accuracy initially and still higher accuracy after learning the user's own records. We compared our method with previous methods that can learn new samples incrementally [20], [21], [56], [16], [2]. We used FLD-213 to train the initial classifier, which was common to all users, and FLD-PERSONAL to evaluate personalization performance. We determined $w = 0.85$ as the weighting value of the SPC in Eq. (4) by 2-fold cross-validation. All the incremental methods used in our comparison used 213 mean vectors as an initial state.

TABLE I
ACCURACY OF VARIOUS FEATURES EXTRACTED FROM GOOGLENET WITH BATCH NORMALIZATION [51] ON FIXED-CLASS CLASSIFICATION USING THE FLD-213 DATASET AND INCREMENTAL CLASSIFICATION USING THE FLD-PERSONAL DATASET. INCREMENTAL CLASSIFICATION RESULTS CORRESPOND TO 1-NN* IN TABLE II.

| feature | dimentionality | fine-tuned | Fixed-class top-1 | Fixed-class top-5 | Incremental top-1 | Incremental top-5 |
|---|---|---|---|---|---|---|
| pool5 [51] | 1024 |  | 25.7 | 52.8 | 32.2 | 46.5 |
| classifier [51] | 1000 |  | 2.3 | 4.9 | 30.7 | 45.0 |
| pool5 [51] | 1024 | ✓ | 52.9 | 80.8 | **38.8** | **53.3** |
| classifier [51] | 213 | ✓ | 49.8 | 77.8 | 37.3 | 52.0 |
| prob [51] | 213 | ✓ | **54.9** | **82.0** | 32.0 | 48.9 |
| lifted [55] | 1024 | ✓ | 43.5 | 71.6 | 29.7 | 45.5 |

The results are shown in Table II. First, the fixed-class CNN [51] showed almost constant performance. Classifier adaptation based on previous data distributions [16], which reranks the output of the CNN classifier, was able to increase accuracy, but it never outperformed the upper limit of the fixed-class classifier because the former cannot learn novel classes. Personalization using cross-entropy (CEL) [2], which combines label frequency in personal data and visual classifiers, also assumes that all classes are known a priori. We modified the frequency-based part of this method to classify novel classes, but it did not perform well. Softratio [56], which optimizes the projection matrix every time a new sample arrives, could not learn from only one sample, and its performance degraded with time. It was not assumed in Softratio that arriving samples may be from existing classes. This result indicates that Softratio could not learn new samples for existing classes. For NCM [20], we tried two experimental settings, $NCM_{800}$ and $NCM_1$. For $NCM_{800}$, all 800 samples used for training the CNN were utilized for mean vector calculations, and accuracy changed very little. This is because the method cannot deal with intra-class diversity. For $NCM_1$, on the other hand, we tried to use only pre-calculated mean vectors for recalculation of mean vectors to accelerate personalization, but this resulted in worse scores. ABACOC [21] can learn incrementally, but 1-NN [20] performed better because ABACOC is affected by the number of samples per class and therefore is not suitable for one-shot learning. We also show two results for 1-NN. "1-NN" has common initial vectors of FLD-213, and "1-NN*" does not have them. A comparison of 1-NN and 1-NN* shows that the initial mean vectors were helpful for solving the cold start problem. On the other hand, the initial mean vectors did not work well at large $t$ because the unnecessary mean vectors for each user continued to affect the classification results. Our proposed SPC significantly outperformed both fixed-class and incremental classifiers at any $t$ by initial mean vectors and weighting parameter $w$. It was shown that the local domain adaptation illustrated in Fig. 6 performs well by the fact that our method beats the 1-NN* method at large $t$.

We show the details of the transition of mean accuracies in Fig. 8. It is clear that our method improves classification accuracy. This demonstrates that the SPC can learn novel concepts from a small number of incremental samples.

We also show the results of two different cases in Table III: the input is within / outside the initial 213 classes. Classification accuracy of initial classes increased as the number of records increased at first, and then became stable. That of user-defined classes started from zero and largely increased as the records are newly acquired by personalization. Detailed examples are shown in Appendix C.

**Effect of the parameter** $w$: Up to this point, we used $w = 0.85$ for the parameter in Eq. (4). In Table IV, we show how the accuracies varied when the parameter value was changed. The accuracies decreased when the parameter value was increased or decreased from $w = 0.85$. The SPC performance is relatively robust to variations in the parameter value. A deviation of about $\pm 0.1$ from the best parameter value resulted in a slight accuracy decline; nevertheless, the resulting accuracies were better than those of previous methods.

**Failure cases**: We showed some failure cases in which the fixed-class CNN [51] could predict the classes correctly but our SPC could not in Fig. 9. The SPC weighted users' inputs more heavily. Thus, when an image has similar visual appearance to the user's previous records of other classes, it tends to be misclassified.

## VI. DISCUSSION

We proposed a reasonable solution for classifier personalization. For each food image registration, our SPC calculate similarity between the input and the vectors in the user's database $V_u$ and the common database $V_m$; therefore, the number of similarity calculation is $213 + t$ after $t$ records. This calculation increases linearly according to the number of records. Therefore, computation required for a person is not large. After two-year operation of our food-logging application, the most heaviest user has registered 9,238 records. If he would record to register food images at this rate for 10 years, the number of records would reach at most 100,000. This number is not so large yet. For more efficient computation, methods such as product quantization or hashing will be applicable.

An important assumption of SPC is that users' records are not very noisy. If a user's records are randomly labeled, SPC does not work. In our previous study on quality of the food records, a registered dietitian evaluated random samples of users who have two weeks long recording and wrong labels were less than 0.5% [57]. Thus, we assume the user labels are not noisy. In addition, a user's records dose not influence the others in our framework. Therefore, even if someone records random labels, they do not affect the performance of recognition of other users.

We think the SPC has still limitations. The SPC can deal with the difference of visual concept between the majority and an individual user by user's incremental inputs. However, the SPC cannot cope with the concept change within a user because it does not have any mechanism to forget old records. How to take newness of records into account is left to the future work.

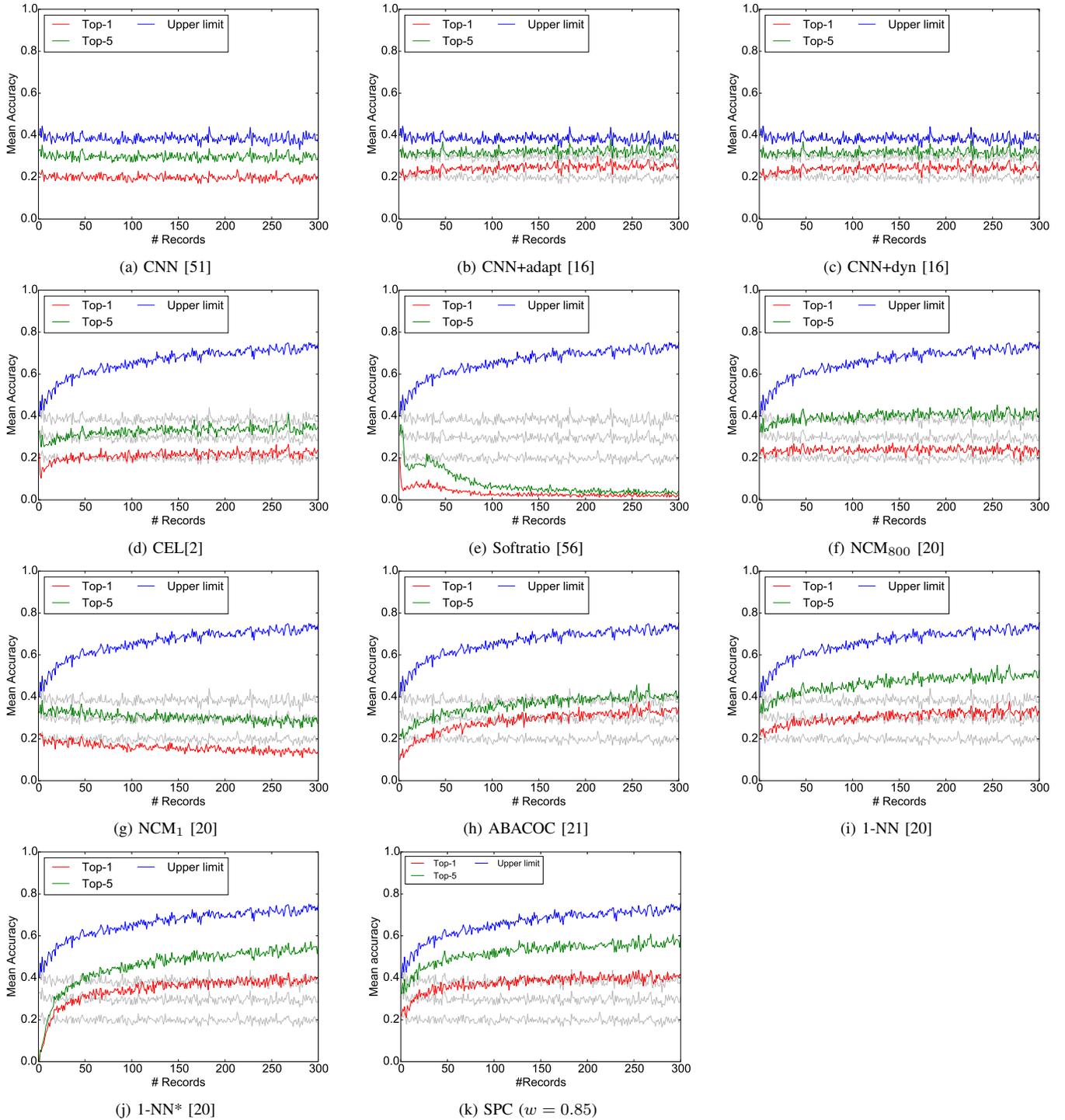

Fig. 8. Time series transition of mean accuracy of each method. We also show top-1 accuracy, top-5 accuracy, and upper limits for the CNN-based fixed-class method in gray lines in each figure.

TABLE II
RESULTS OF PREVIOUS METHODS AND OUR SEQUENTIAL PERSONALIZED CLASSIFIER (SPC) ON PERSONALIZED FOOD RECOGNITION. EACH CELL SHOWS THE AVERAGE MEAN ACCURACY OVER 50 CONSECUTIVE IMAGES IN ORDER BY TIME. FEAT IDENTIFIES THE FEATURES EXTRACTED FROM GOOGLENET WITH BATCH NORMALIZATION [51] THAT WERE USED. A ✓ IN THE NVL COLUMN INDICATES THAT THE METHOD COULD LEARN NOVEL CLASSES. THE TWO UPPER LIMIT ROWS SHOW THE RATES AT WHICH THE CLASS OF THE ARRIVING SAMPLE WAS IN $V_M$ OR $V_M \cup V_u$, RESPECTIVELY. $V_M$ IS THE SET OF THE INITIAL MEAN VECTORS AND $V_u$ IS THE SET OF USER $u$'S INPUTS.

| Method | FEAT | NVL | $t_1$–$t_{50}$ top-1 | top-5 | $t_{51}$–$t_{100}$ top-1 | top-5 | $t_{101}$–$t_{150}$ top-1 | top-5 | $t_{151}$–$t_{200}$ top-1 | top-5 | $t_{201}$–$t_{250}$ top-1 | top-5 | $t_{251}$–$t_{300}$ top-1 | top-5 |
|---|---|---|---|---|---|---|---|---|---|---|---|---|---|---|
| CNN [51] | prob | | 19.9 | 30.3 | 19.8 | 29.3 | 19.6 | 29.6 | 19.4 | 29.5 | 19.7 | 29.6 | 19.5 | 29.4 |
| CNN+adapt [16] | prob | | 22.5 | 31.5 | 23.9 | 31.3 | 24.5 | 32.1 | 24.9 | 32.2 | 25.1 | 32.5 | 25.3 | 32.4 |
| CNN+dyn [16] | prob | | 22.5 | 31.5 | 23.9 | 31.3 | 24.4 | 31.9 | 24.3 | 31.9 | 24.4 | 32.0 | 24.3 | 31.8 |
| CEL [2] | prob | ✓ | 18.5 | 29.5 | 20.8 | 32.1 | 21.6 | 33.1 | 21.9 | 33.6 | 22.2 | 33.9 | 22.6 | 34.2 |
| Softratio [56] | pool5 | ✓ | 7.4 | 19.1 | 3.9 | 9.7 | 2.7 | 6.1 | 2.4 | 4.7 | 2.2 | 4.1 | 2.1 | 3.7 |
| NCM$_{800}$ [20] | pool5 | ✓ | 23.3 | 37.1 | 23.8 | 39.1 | 23.9 | 40.3 | 23.9 | 40.7 | 24.0 | 41.0 | 23.7 | 41.0 |
| NCM$_1$ [20] | pool5 | ✓ | 18.7 | 33.1 | 16.7 | 31.3 | 15.9 | 30.5 | 15.2 | 30.3 | 14.5 | 28.9 | 13.8 | 28.4 |
| ABACOC [21] | pool5 | ✓ | 19.2 | 27.6 | 26.3 | 33.7 | 29.1 | 36.6 | 31.1 | 38.5 | 32.0 | 39.4 | 33.3 | 40.7 |
| 1-NN [20] | pool5 | ✓ | 25.8 | 39.4 | 28.7 | 44.1 | 30.6 | 46.8 | 31.8 | 48.5 | 32.5 | 49.3 | 32.9 | 50.3 |
| 1-NN* [20] | pool5 | ✓ | 23.1 | 29.0 | 33.0 | 43.2 | 35.9 | 47.8 | 37.4 | 50.0 | 38.1 | 51.7 | 38.8 | 53.3 |
| **SPC** ($w=0.85$) | pool5 | ✓ | **31.4** | **43.5** | **36.5** | **50.5** | **38.6** | **53.4** | **40.6** | **54.8** | **39.7** | **55.4** | **40.2** | **56.6** |
| Upper limit | – | | 39.1 | | 38.0 | | 38.4 | | 38.3 | | 38.2 | | 38.1 | |
| Upper limit | – | ✓ | 54.6 | | 62.8 | | 66.9 | | 69.5 | | 70.9 | | 72.4 | |

TABLE III
BREAKDOWN OF THE RESULT OF OUR SEQUENTIAL PERSONALIZED CLASSIFIER (SPC). MEAN OF CONDITIONAL ACCURACIES WHEN IMAGES ARE WITHIN THE INITIAL 213 CLASSED AND OUTSIDE THE INITIAL 213 CLASSES ARE SHOWN.

| | $t_1$–$t_{50}$ top-1 | top-5 | $t_{51}$–$t_{100}$ top-1 | top-5 | $t_{101}$–$t_{150}$ top-1 | top-5 | $t_{151}$–$t_{200}$ top-1 | top-5 | $t_{201}$–$t_{250}$ top-1 | top-5 | $t_{251}$–$t_{300}$ top-1 | top-5 |
|---|---|---|---|---|---|---|---|---|---|---|---|---|
| Images from initial 213 classes | 55.1 | 79.6 | 57.9 | 81.2 | 58.4 | 81.1 | 58.6 | 80.5 | 58.4 | 80.5 | 58.4 | 80.9 |
| Images not from initial 213 classes | 16.1 | 20.4 | 23.3 | 31.6 | 26.2 | 36.2 | 27.8 | 38.8 | 28.2 | 40.0 | 29.0 | 41.5 |

TABLE IV
EFFECT OF THE VALUE OF THE WEIGHTING PARAMETER $w$ BETWEEN THE FIXED-CLASS COMMON CLASSIFIER AND THE USER-SPECIFIC INCREMENTAL CLASSIFIER ON THE ACCURACY OF OUR SEQUENTIAL PERSONALIZED CLASSIFIER (SPC).

| $w$ | $t_1$–$t_{50}$ top-1 | top-5 | $t_{51}$–$t_{100}$ top-1 | top-5 | $t_{101}$–$t_{150}$ top-1 | top-5 | $t_{151}$–$t_{200}$ top-1 | top-5 | $t_{201}$–$t_{250}$ top-1 | top-5 | $t_{251}$–$t_{300}$ top-1 | top-5 |
|---|---|---|---|---|---|---|---|---|---|---|---|---|
| 0.70 | 28.5 | 42.5 | 34.0 | 47.8 | 36.3 | 50.6 | 37.8 | 52.0 | 38.4 | 53.0 | 38.9 | 54.2 |
| 0.75 | 30.1 | 43.6 | 34.9 | 49.4 | 37.0 | 52.0 | 38.2 | 53.2 | 38.7 | 54.0 | 39.2 | 55.1 |
| 0.80 | 31.1 | **43.9** | 35.9 | 50.3 | 37.9 | 53.1 | 38.9 | 54.4 | 39.1 | 54.9 | 39.7 | 56.1 |
| 0.85 | **31.4** | 43.5 | **36.5** | **50.5** | **38.6** | **53.4** | **40.6** | **54.8** | **39.7** | **55.4** | **40.2** | **56.6** |
| 0.90 | 30.4 | 42.6 | 35.8 | 49.2 | 38.0 | 52.3 | 39.3 | 54.0 | **39.7** | 54.8 | **40.2** | 55.9 |
| 0.95 | 28.3 | 41.1 | 32.9 | 46.9 | 35.3 | 50.0 | 36.6 | 51.8 | 37.4 | 52.8 | 37.9 | 53.7 |
| 1.00 (1-NN [20]) | 25.8 | 39.4 | 28.7 | 44.1 | 30.6 | 46.8 | 31.8 | 48.5 | 32.5 | 49.3 | 32.9 | 50.3 |

Our SPC also has difficulty on updating the classification model, that is, the architecture or weights of the convolutional neural network. When using a fixed-class classifier, only we have to do is replace the model to a new one. On the other hand, when using our SPC, we have to extract feature vectors from all the previous inputs and reconstruct $V_u$ and $V_m$ to replace the feature extractor. Although this process can be done in background, the calculation cost is high especially when the number of records is large. Updating classifiers is a further issue.

Using hierarchies of food classes is also remaining as a challenging problem, as described in Section III. We cannot apply hierarchical classification for FLD because the labels defined by users do not have hierarchies. However, doing extra annotations of the user-defined labels and using the hierarchy may be beneficial to improve classification accuracy.

## VII. CONCLUSION

In this paper, we have introduced a personalization problem in food image recognition and proposed a method for classifier personalization. Personalized food recognition contains problems of incremental learning, domain adaptation, and one-shot learning. Our method combines a fixed-class NCM classifier that performs competitively with CNN and a user-specific classifier that learns the user's input incrementally, with heavier weighting given to later input. In contrast to existing studies in which experiments were conducted using artificial scenarios, we introduced a new dataset to evaluate personalized classification performance in realistic situations. Our proposed SPC significantly outperformed the existing

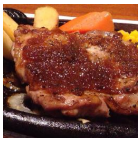

$t = 22$
Ground truth: *Chicken steak*

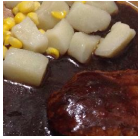

$t = 39$
Predict: *Chicken steak*
Ground truth: *Demi-glace hamburger steak*

(a) User A

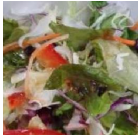

$t = 46$
Ground truth: *Mixed salad*

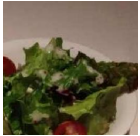

$t = 47$
Predict: *Mixed salad*
Ground truth: *Caesar salad*

(b) User B

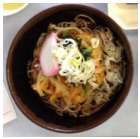

$t = 9$
Ground truth: *Soba with mixed tempura*

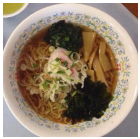

$t = 29$
Predict: *Soba with mixed tempura*
Ground truth: *Soy sauce ramen*

(c) User C

Fig. 9. Failure cases of our sequential personalized classifier (SPC). The images on the right are misclassified as the classes of their left-hand images. $t$ shows the time when the image was input.

methods in terms of accuracy. In future work, we plan to accelerate this work on personalization by using information from other users whose labeling tendencies are similar to that of the target user.

## APPENDIX A
## LABELS IN FLD-213.

We show the labels included in FLD-213 in descending order of their number of records.

*Rice, Green salad, Natto, Yogurt, Black coffee, Miso soup (seaweed & tofu), Banana, Cold tofu, Shredded cabbage, Fresh vegetable salad, Plain yogurt, Wiener, Miso soup (seaweed & green onion), Fried chicken, Apple, Japanese omelet, Boiled egg, Mini tomato, Eggs fried sunny-side up, Tomato, Coffee, Potato salad, Vegetable juice, Buttered toast, Glass of milk, Broccoli salad, Boiled spinach, Cafe au lait, Meat and vegetable stir-fry, Bread (sliced into six pieces), Pan-fried dumpling, Simmered hijiki seaweed, Baked salmon, Mandarin orange, Tomato salad, Mixed salad, Vegetable stir-fry, Brown rice, Pork miso soup, Braised pumpkin, Miso soup (Japanese radish), Miso soup (tofu & green onion), Kimchi, Ginger-fried pork, Canned beer (350 mL), Meat and potato stew, French fries, Yellow pickled radish, Macaroni salad, Rice with mixed grains, Sauce chow mein, Strawberry, Cucumber, Five-ingredient rice, Cup of green tea, Kinpira gobo, Miso soup (Japanese radish & deep-fried tofu), Potato croquette, Coleslaw, Pork curry, Soy milk, Soy sauce ramen, Green soybeans, Roasted mackerel, French bread, Bread roll, Japanese radish salad, Chinese soup, Draft beer, Minestrone, Beam sprout stir-fry, Distilled spirits, Mapo tofu, Demi-glace hamburger steak, Chicken teriyaki, Beef curry, Scrambled eggs, Pork dumplings, Roast ham, Red wine, Miso soup (potato & onion), Chicken curry, Chicken sauté, Oden, Rice ball (salmon), Raw egg, Japanese salad, Sliced raw tuna, Boiled barley and rice, Dried radish strips, Three sorts of sliced raw fishes, Kiwi, Vinegared mozuku seaweed, Simmered root vegetables with chicken, Persimmon, Zaru soba, Cream stew, Miso soup (cabbage & deep-fried tofu), Pork ramen, Bread (sliced into eight pieces), Miso soup (spinach & deep-fried tofu), Spicy sautéed lotus root, Canned low-malt beer (350 mL), Mince cutlet, Salted Chinese cabbage, Thick omelet, Beef bowl, Barley tea, Spinach with sesame dressing, Vinegared cucumber, Pork loin cutlet, Spaghetti Bolognese, Braised eggplant, Sliced onion, Pear, Mackerel simmered in miso, Miso soup (seaweed, green onion, & wheat gluten), Miso soup (freshwater clam), Caesar salad, Seaweed soup, Consommé, Omelet rice, Boiled and roasted tofu refuse, Boiled Japanese mustard spinach, Iced coffee, Cereal with milk, Milk tea, Japanese radish (oden), Spaghetti Napolitana, Grapefruit, Avocado, Japanese hamburger steak, Kake udon, Pickled plum, Rice ball (pickled plum), Liquid yogurt, Mixed fried rice, Salted salmon, Vermicelli soup, Sushi wrapped in fried tofu, Bowl of rice topped with chicken and eggs, Spring roll, Grilled deep-fried tofu, Boiled soybeans, Ham and eggs, Rice ball (kelp), Black tea, Rice ball (salt), Chicken cutlet, Grilled saury, Yellowtail teriyaki, Assorted hand-rolled sushi, Egg custard, Chicken salad, Assorted sandwiches, Instant corn cream soup, Hamburger sandwich, Tuna salad, Boiled okra, Instant noodles, Bacon and eggs, Adjusted soy milk, Straight tea, Grilled eggplant, Pot-au-feu, Mushroom soup, Fried white fish, Simmered taro, Steamed vegetables, Croissant, Japanese omelet, Gigantic peak, Pineapple, Orange, Ham sandwich, Assorted tempura, Fried fish ball, Pork sauté, Glass of low-fat milk, Hot dog, Boiled tofu, Kanto-style rolled omelet, Small chocolate, Steamed meat bun, Toast and jelly, Seasoned cod roe, Roast pork, Onion soup, Rice porridge with egg, Miso soup (nameko mushroom), Cold vermicellifine noodles, Seaweed salad, Sukiyaki, Miso soup (tofu & nameko mushroom), Kimchi casserole, Fillet pork cutlet, Hot spring egg, Banana yogurt smoothie, Miso ramen, Octopus dumplings, Scrambled eggs, Assorted fruits, Sautéed spinach, Short rib, Rice with raw egg, Cafe latte (short size), Plain omelet, Chicken nugget, Fried shrimp with tartar sauce, Pork cullet bowl, Beer (500 mL), Sliced cheese, Grated radish.*

APPENDIX B
COMBINING STRATEGY OF THE FIXED-CLASS CLASSIFIER AND THE USER-SPECIFIC CLASSIFIER

We used Eq. (4) to obtain classification results of our SPC. As described in Section IV, we also tried a linear combination of the fixed-class classifier and the user-specific classieir as in Eq. (5): we call this classifier $SPC_{sum}$. However, as shown in Table V, $SPC_{sum}$ did not work well. As increasing the parameter $w_s$, the accuracy got worse. This is because $SPC_{sum}$ is more likely to return the classes in both $C_u$ and $C_m$.

APPENDIX C
USER EXPERIENCE

Our purpose of this paper is to help the labeling step in the food logging application [50]. In this section, we show the close-up results of the fixed-class classifier and the SPC to reproduce users' experiences. Table VI shows the 1st, 2nd, 3rd, 298th, 299th, and 300th records of the three users. The users are supposed to select the appropriate labels if it exists in the label candidates provided by the algorithm. The SPC presents the correct labels more frequently than the fixed class CNN, and it is expected to reduce the time and effort to enter the correct label by taking user-defined labels into account incrementally.

ACKNOWLEDGMENT

This work was partially supported by JST CREST Grant Number JPMJCR1686 and JSPS KAKENHI Grant Number 18H03254.

TABLE V
THE RESULTS OF $\mathrm{SPC}_{sum}$, A LINEAR COMBINATION OF THE COMMON AND THE USER-SPECIFIC CLASSIFIER. $w_s$ IS THE WEIGHTING COEFFICIENT.

| $w_s$ | $t_1$–$t_{50}$ | | $t_{51}$–$t_{100}$ | | $t_{101}$–$t_{150}$ | | $t_{151}$–$t_{200}$ | | $t_{201}$–$t_{250}$ | | $t_{251}$–$t_{300}$ | |
|---|---|---|---|---|---|---|---|---|---|---|---|---|
| | top-1 | top-5 | top-1 | top-5 | top-1 | top-5 | top-1 | top-5 | top-1 | top-5 | top-1 | top-5 |
| 0.0 (1-NN* [20]) | 23.1 | 29.0 | 33.0 | 43.2 | 35.9 | 47.8 | 37.4 | 50.0 | 38.1 | 51.7 | 38.8 | 53.3 |
| 0.1 | 23.2 | 31.9 | 32.3 | 43.0 | 34.6 | 47.7 | 36.0 | 49.6 | 36.3 | 51.1 | 36.9 | 52.5 |
| 0.2 | 21.6 | 31.2 | 29.0 | 41.1 | 30.9 | 44.8 | 31.4 | 45.9 | 31.6 | 46.8 | 31.9 | 47.9 |
| 0.3 | 19.2 | 29.7 | 25.2 | 36.8 | 26.6 | 39.4 | 27.0 | 39.7 | 27.4 | 40.3 | 27.7 | 40.9 |
| 0.4 | 17.5 | 29.5 | 22.3 | 31.7 | 23.7 | 33.3 | 24.2 | 33.3 | 24.9 | 33.8 | 25.0 | 34.2 |
| 0.5 | 16.3 | 27.5 | 21.0 | 27.4 | 22.8 | 29.1 | 23.2 | 29.3 | 24.1 | 30.2 | 24.5 | 30.5 |
| 0.6 | 17.3 | 27.1 | 21.0 | 26.4 | 22.8 | 27.8 | 23.1 | 28.2 | 24.0 | 29.0 | 24.4 | 29.5 |
| 0.7 | 20.3 | 30.4 | 22.2 | 29.6 | 23.4 | 30.2 | 23.6 | 30.1 | 24.3 | 30.5 | 24.5 | 30.7 |
| 0.8 | 22.7 | 31.8 | 23.8 | 31.5 | 24.6 | 32.0 | 24.5 | 31.8 | 24.9 | 31.9 | 24.9 | 31.9 |
| 0.9 | 23.3 | 31.6 | 24.4 | 31.5 | 24.8 | 32.0 | 24.6 | 31.9 | 24.9 | 32.0 | 24.7 | 32.0 |
| 1.0 | 19.8 | 30.0 | 19.3 | 28.9 | 19.3 | 29.1 | 19.1 | 28.9 | 19.4 | 29.2 | 19.1 | 28.9 |

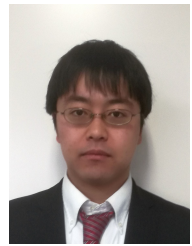

**Shota Horiguchi** received B.S. in Information and Communication Engineering and M.S. in Information Science and Technology from the University of Tokyo in 2015 and 2017, respectively. He is currently a researcher at Hitachi, Ltd.

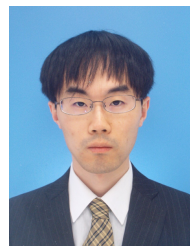

**Sosuke Amano** received BS in Infomation and Communication Eng. and MS in Interdiciplinary Studies in Information Science at the University of Tokyo in 2012 and 2015 respectively. He is a Ph.D student of Dept,. of Information and Commnication Eng. He also joined foo.log Inc. in 2015.

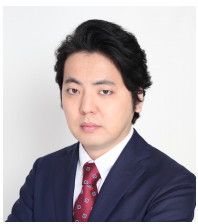

**Makoto Ogawa** received B.S. in Electronic Engineering, M.S. and Ph.D. degree in Science from the University of Tokyo, in 2000, 2002, and 2005, respectively. He is currently a CEO of foo.log Inc., Tokyo, which runs multimedia food recording service, FoodLog.

TABLE VI
Close-up results of the fixed-class CNN classifier [51] and our sequential personalized classifier (SPC). For each user, the 1st, 2nd, and 3rd records are desplayed in the upper row, and the 298th, 299th, and 300th records are displayed in the lower row. The user-defined labels are written in *ITALIC* letters, and the correct labels are written in **BOLD** letters.

(a) User A

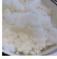
$t = 1$
*Rice*

| Rank | CNN [51] | SPC |
|---|---|---|
| 1 | **Rice** | **Rice** |
| 2 | Boiled barley and rice | Boiled barley and rice |
| 3 | Rice with raw egg | Rice with raw egg |
| 4 | Brown rice | Brown rice |
| 5 | Chicken curry | Chicken curry |

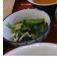
$t = 2$
*Stewed fried tofu and Japanese mustard spinach*

| Rank | CNN [51] | SPC |
|---|---|---|
| 1 | Boiled Japanese mustard spinach | Boiled Japanese mustard spinach |
| 2 | Boiled spinach | Boiled spinach |
| 3 | Spinach with sesame dressing | Spinach with sesame dressing |
| 4 | Sautéed spinach | Sautéed spinach |
| 5 | Boiled okra | Boiled okra |

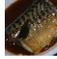
$t = 3$
*Mackerel simmered in miso*

| Rank | CNN [51] | SPC |
|---|---|---|
| 1 | **Mackerel simmered in miso** | **Mackerel simmered in miso** |
| 2 | Roasted mackerel | Roasted mackerel |
| 3 | Yellowtail teriyaki | Yellowtail teriyaki |
| 4 | Grilled saury | Grilled saury |
| 5 | Roast pork | Roast pork |

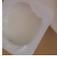
$t = 298$
*Danone Bio Yogurt*

| Rank | CNN [51] | SPC |
|---|---|---|
| 1 | Pear | ***Danone Bio Yogurt*** |
| 2 | Boiled egg | Yogurt |
| 3 | Glass of milk | Glass of milk |
| 4 | Japanese radish (oden) | Adjusted soy milk |
| 5 | Plain yogurt | Cold tofu |

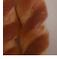
$t = 299$
*Butter stick*

| Rank | CNN [51] | SPC |
|---|---|---|
| 1 | Bread roll | ***Butter stick*** |
| 2 | Croissant | Butter-enriched roll |
| 3 | Hot dog | Mini croissant |
| 4 | French bread | Bread roll |
| 5 | Pickled plum | Croissant |

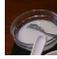
$t = 300$
*Almond jelly*

| Rank | CNN [51] | SPC |
|---|---|---|
| 1 | Plain yogurt | ***Almond jelly*** |
| 2 | Glass of milk | Danone Bio Yogurt |
| 3 | Yogurt | Yogurt |
| 4 | Glass of low-fat milk | Glass of milk |
| 5 | Liquid yogurt | Egg custard |

(c) User B

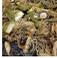
$t = 1$
*Mongolian mutton barbecue*

| Rank | CNN [51] | SPC |
|---|---|---|
| 1 | Sauce chow mein | Sauce chow mein |
| 2 | Sukiyaki | Sukiyaki |
| 3 | Meat and vegetable stir-fry | Meat and vegetable stir-fry |
| 4 | Vegetable stir-fry | Vegetable stir-fry |
| 5 | Short rib | Short rib |

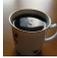
$t = 2$
*Black coffee*

| Rank | CNN [51] | SPC |
|---|---|---|
| 1 | **Black coffee** | **Black coffee** |
| 2 | Coffee | Coffee |
| 3 | Cafe au lait | Cafe au lait |
| 4 | Straight tea | Straight tea |
| 5 | Iced coffee | Iced coffee |

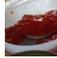
$t = 3$
*Salmon roe*

| Rank | CNN [51] | SPC |
|---|---|---|
| 1 | Chicken sauté | Chicken sauté |
| 2 | Pork sauté | Pork sauté |
| 3 | Chicken teriyaki | Chicken teriyaki |
| 4 | Demi-glace hamburger steak | Demi-glace hamburger steak |
| 5 | Seasoned cod roe | Seasoned cod roe |

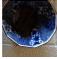
$t = 298$
*Kelp*

| Rank | CNN [51] | SPC |
|---|---|---|
| 1 | Small chocolate | ***Kelp*** |
| 2 | Vinegared mozuku seaweed | *Yogurt with pomegranate vinegar* |
| 3 | Mackerel simmered in miso | *Small chocolate* |
| 4 | Rice ball (kelp) | *Chocolanthanum* |
| 5 | Rice ball (salmon) | *Pickled herring* |

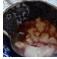
$t = 299$
*Yogurt with pomegranate vinegar*

| Rank | CNN [51] | SPC |
|---|---|---|
| 1 | Egg custard | ***Yogurt with pomegranate vinegar*** |
| 2 | Boiled tofu | *Boiled scallops* |
| 3 | Cold tofu | *Egg custard* |
| 4 | Plain yogurt | *Grilled bacon* |
| 5 | Onion soup | *Pickled herring* |

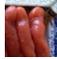
$t = 300$
*Cod roe*

| Rank | CNN [51] | SPC |
|---|---|---|
| 1 | Wiener | **Cod roe** |
| 2 | Seasoned cod roe | Grilled cod roe |
| 3 | Sliced raw tuna | Wiener |
| 4 | Pickled plum | Seasoned cod roe |
| 5 | Hot dog | *Now Hamburger* |

(e) User C

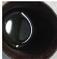
$t = 1$
*Glass of green tea*

| Rank | CNN [51] | SPC |
|---|---|---|
| 1 | Black coffee | Black coffee |
| 2 | Barley tea | Barley tea |
| 3 | Black tea | Black tea |
| 4 | Distilled spirits | Distilled spirits |
| 5 | Coffee | Coffee |

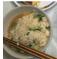
$t = 2$
*Rice*

| Rank | CNN [51] | SPC |
|---|---|---|
| 1 | Rice porridge with egg | Rice porridge with egg |
| 2 | Boiled and roasted tofu refuse | Boiled and roasted tofu refuse |
| 3 | Five-ingredient rice | Five-ingredient rice |
| 4 | Brown rice | Brown rice |
| 5 | Mixed fried rice | Mixed fried rice |

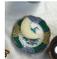
$t = 3$
*Sliced onion*

| Rank | CNN [51] | SPC |
|---|---|---|
| 1 | Yogurt | Yogurt |
| 2 | Plain yogurt | Plain yogurt |
| 3 | Egg custard | Egg custard |
| 4 | Apple | Apple |
| 5 | Pear | Pear |

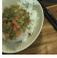
$t = 298$
*Rish bowl topped with deep-fried fish*

| Rank | CNN [51] | SPC |
|---|---|---|
| 1 | Rice | Rice |
| 2 | Boiled barley and rice | Black coffee |
| 3 | Brown rice | ***Rish bowl topped with deep-fried fish*** |
| 4 | Natto | Tuna salad |
| 5 | Rice with mixed grains | Natto |

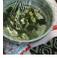
$t = 299$
*Stewed soy beans and kelp*

| Rank | CNN [51] | SPC |
|---|---|---|
| 1 | Seaweed soup | ***Stewed soy beans and kelp*** |
| 2 | Chinese soup | *Miso soup (bean sprouts)* |
| 3 | Vermicelli soup | *Seaweed Udon* |
| 4 | Boiled tofu | *Salted Chinese cabbage* |
| 5 | Seaweed salad | *Udon Hot Pot* |

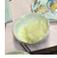
$t = 300$
*Fish and vegetables seasoned in vinegar*

| Rank | CNN [51] | SPC |
|---|---|---|
| 1 | Sliced cheese | Kanto-style rolled omelet |
| 2 | Salted Chinese cabbage | Apple |
| 3 | Apple | Japanese radish in soy sauce |
| 4 | Avocado | ***Fish and vegetables seasoned in vinegar*** |
| 5 | Cold tofu | Yellow pickled radish |

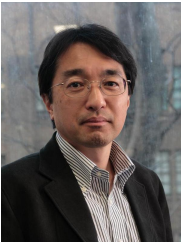

**Kiyoharu Aizawa** (F'2016) received the B.E., the M.E., and the Dr.Eng. degrees in Electrical Engineering all from the University of Tokyo, in 1983, 1985, 1988, respectively. He is currently a Professor at Department of Information and Communication Engineering of the University of Tokyo. He was a Visiting Assistant Professor at University of Illinois from 1990 to 1992. His research interest is in image processing and multimedia applications. He received the 1987 Young Engineer Award and the 1990, 1998 Best Paper Awards, the 1991 Achievement Award, 1999 Electronics Society Award from IEICE Japan, and the 1998 Fujio Frontier Award, the 2002 and 2009 Best Paper Award, and 2013 Achievement award from ITE Japan. He received the IBM Japan Science Prize in 2002. He is currently a Senior Associate Editor of IEEE Tras. Image Processing, and on Editorial Board of ACM TOMM, APSIPA Transactions on Signal and Information Processing, and International Journal of Multimedia Information Retrieval. He served as the Editor in Chief of Journal of ITE Japan, an Associate Editor of IEEE Trans. Image Processing, IEEE Trans. CSVT and IEEE Trans. Multimedia. He has served a number of international and domestic conferences; he was a General co-Chair of ACM Multimedia 2012. He is a council member of Science Council of Japan.